\documentclass[journal]{IEEEtran}

\ifCLASSINFOpdf
  \usepackage[pdftex]{graphicx}
\else
  \usepackage[dvips]{graphicx}
\fi
\usepackage{amsmath}
\usepackage{amsfonts}
\usepackage{amssymb}
\usepackage{amsthm}
\usepackage{algorithmic}
\usepackage[export]{adjustbox}

\usepackage{array}
\usepackage{tabularx}
\usepackage{multirow}
\usepackage{rotating}
\usepackage{array}

\usepackage[caption=false,font=footnotesize]{subfig}

\usepackage{changepage}

\usepackage{booktabs}

\begin{document}
%
\pagenumbering{gobble}

\title{Deep Learning Computed Tomography based on the Defrise and Clack Algorithm}

\author{Chengze~Ye,
        Linda-Sophie~Schneider,
        Yipeng~Sun,
        and~Andreas~Maier
\thanks{All authors are with the Pattern Recognition Lab, Friedrich-Alexander University Erlangen-Nuremberg, Erlangen, Germany (e-mail: chengze.ye@fau.de)}
}

\maketitle

\begin{abstract}

This study presents a novel approach for reconstructing cone beam computed tomography (CBCT) for specific orbits using known operator learning. Unlike traditional methods, this technique employs a filtered backprojection type (FBP-type) algorithm, which integrates a unique, adaptive filtering process. This process involves a series of operations, including weightings, differentiations, the 2D Radon transform, and backprojection. The filter is designed for a specific orbit geometry and is obtained using a data-driven approach based on deep learning. The approach efficiently learns and optimizes the orbit-related component of the filter. The method has demonstrated its ability through experimentation by successfully learning parameters from circular orbit projection data. Subsequently, the optimized parameters are used to reconstruct images, resulting in outcomes that closely resemble the analytical solution. This demonstrates the potential of the method to learn appropriate parameters from any specific orbit projection data and achieve reconstruction. The algorithm has demonstrated improvement, particularly in enhancing reconstruction speed and reducing memory usage for handling specific orbit reconstruction.
\end{abstract}


\IEEEpeerreviewmaketitle

\section{Introduction}


\IEEEPARstart{C}{omputed} tomography (CT) is a medical imaging technique that uses X-ray imaging to produce high-resolution sectional images of the human body. It is essential for medical diagnosis, surgical planning, and scientific research. CT technology is also crucial in the industrial sector, especially in non-destructive testing, allowing engineers to detect and analyze minor internal defects in materials, ensuring quality and safety.

Traditional CT scans usually involve a rotating structure with fixed orbits for the X-ray source and detector. In contrast, recent C-arm CT systems provide greater flexibility by allowing free movement of these components. Flexible CT orbits are especially advantageous in interventional imaging, where they help to avoid metal artifacts and optimize image acquisition speed and radiation dose \cite{orbit}. In the industrial sector, this approach speeds up inspections, improving the efficiency of identifying and analyzing structural defects or anomalies. However, if the orbit of the X-ray source follows a specific non-circular path, this presents a challenge to the reconstruction process.

Previous research has proposed various methods for reconstructing specific orbits. For example, Zeng \cite{zeng} described the algebraic iterative reconstruction algorithm, which can reconstruct cone beam computed tomography (CBCT) projections of arbitrary orbits.

Grangeat \cite{Grangeat} proposed a reconstruction method based on the relation between cone-beam data and a function associated with the 3D Radon transform of the images. This algorithm can accurately reconstruct any CT orbit. However, in practical applications, the algorithm requires the prior construction of a matrix containing samples of the intermediate function, which is then subjected to subsequent processing. 

The reconstruction algorithm mentioned above can successfully reconstruct any given orbit. However, it has a slow reconstruction speed and high memory consumption.
To overcome this limitation, Defrise and Clack \cite{Defrise} introduced a filtered backprojection type (FBP-type) algorithm based on the Grangeat method. The algorithm is customized for different orbits by designing specific redundancy weights, ultimately achieving analytic reconstruction for any orbit data.

Calculating specific redundancy weights for different orbits in practical applications can be challenging. However, this problem can be approximated and solved through deep learning techniques. While deep learning is adept at modeling complex functions, it may lead to overly complex models with excessive parameters when dealing with intricate problems such as the inverse problem in CT reconstruction.

Maier \textit{et al.} \cite{maier2019learning} proposed a method to incorporate known operators into machine learning algorithms. This approach significantly reduces the number of neural network parameters, enabling faster training and reducing maximum error bounds.
 

Syben \textit{et al.} \cite{syben2017precision} proposed a neural network based approach that utilizes known operators in the filtered backprojection (FBP) framework. 
It demonstrates the feasibility and effectiveness of optimizing parameters in the specific layer of the reconstruction pipeline.


We present a data-driven methodology for reconstructing CBCT projections for a specific orbit. The method belongs to the category of FBP algorithms and incorporates a sequence of elementary 2D operations as a filter, including weightings, differentiations, 2D projection, and 2D backprojection. A reconstruction framework is established using known operators, where only the redundancy weight associated with the CT orbit are treated as trainable parameters. They are obtained through a training process, where parameters are fitted based on a given orbit, ultimately enabling rapid reconstruction of specific orbit CBCT scans.

\section{Methods}
The principle of CT is based on measuring the absorption of X-rays at different angles. Algebraic algorithms are then used to transform this projection data into an image of the object's interior. 
\subsection{Filtered Backprojection}
Algebraic reconstruction involves solving  $p = Ax$ to recover image or volume $x$ from projection data $p$. The relationship between projection and image domain is captured by matrix $A$. However, computing the inverse of $A$ directly is challenging. To address this, the filtered backprojection algorithm is used \cite{zeng}.

To reconstruct the original object from its projection data $p(s, \theta)$ in a parallel beam geometry, a convolution with a Ramp filter $h(s)$ should be first performed. The goal of this process is to filter the projection data in order to enhance high-frequency components while suppressing lower-frequency components.

\[q(s, \theta) = h(s) \ast p(s, \theta) \tag{1}\]
\[h(s) = \int | \omega | e^{2\pi \omega s} d\omega \tag{2}\]

The filtered data $q(s, \theta)$ is then subjected to a backprojection operation. This step maps the filtered data back into the image space from different angles.
\[f(x, y) = \int_{0}^{\pi} q(s, \theta)\bigg|_{s=x\cdot\cos \theta+y\cdot\sin \theta} d\theta \tag{3}\]

Finally, the reconstructed image or volume $f(x, y)$ can be obtained through this comprehensive procedure.

When using cone beam geometry for projection data, it is necessary to introduce cosine weight to correct the geometry.

\subsection{Grangeat's Inversion}

Grangeat \cite{Grangeat} introduced a reconstruction method that established a connection between cone-beam projections and the first derivative of the Radon transform by identifying an intermediate function. This method achieves rebinning from the Cartesian coordinate system of cone-beam geometry to the spherical coordinate system of the Radon domain using a plane as an information vector. Zeng \cite{zeng} describes a process where numerous lines are sampled on a two-dimensional cone-beam detector. The projection is integrated along these lines, resulting in a line integral that can be viewed as a plane integral weighted by $\frac{1}{r}$. This plane is characterized by the distance $l\in[-B, +B]$ to the origin and a vector $\theta$ orthogonal to the plane. Furthermore, it has been observed that the derivative of a plane integral along the tangential direction $dt$ is equivalent to the derivative of a $\frac{1}{r}$ weighted plane integral along $d\alpha$. Based on this observation, Grangeat's intermediate function~\eqref{eq:myequation2} is derived.
\begin{align*}
S(\theta, \lambda )=-\int_{s^2}^{} d\beta \delta'(\beta, \theta)g(\beta,\lambda)\qquad\theta\in S^2, \lambda\in \Lambda \tag{4}
\label{eq:myequation2}
\end{align*}

Equation~\eqref{eq:myequation2} uses $\lambda\in \Lambda$ as a parameter for the source position $a(\lambda)$ and $\beta\in S^2$ to indicate the direction of the line integral. The function $g(\beta,\lambda)$ represents the cone-beam projection, while $\delta'$ signifies the derivative of the Dirac delta distribution. By incorporating the intermediate function into the inverse Radon transform formula, one can derive the reconstruction formula~\eqref{eq:myequation3000} for Grangeat's method.
\begin{align*}
f(x)=-\frac{1}{4\pi^2}\int_{s^2/2}^{}d\theta\int_{-B}^{B}dl \delta'(x\cdot\theta-l)S(\theta, \lambda(l,\theta) )\tag{5}
\label{eq:myequation3000}
\end{align*}

\subsection{Defrise and Clack Algorithm}
This algorithm is a shift-variant filtered backprojection algorithm derived from Grangeat's method. It can be used to reconstruct projection data of general non-circular orbits. Unlike the Grangeat algorithm, it eliminates the need to store all intermediate functions in a matrix. This enables the projection acquisition and processing to be executed synchronously, which greatly enhances the speed of the reconstruction process. The reconstruction formula is as follows:
\begin{align*}
f(x)=&\int_{ \Lambda}^{} d\lambda\int_{S^2/2}^{}d\theta-\frac{1}{4\pi^2}\mid a'(\lambda)\cdot\theta\mid\frac{1}{n(\theta, \lambda)}\\
&\times\delta' ((x-a(\lambda))\cdot \theta)S(\theta, \lambda ).\tag{6}
\label{eq:myequation3}
\end{align*}

The term $n(\theta, \lambda)$ is defined as the number of intersections between orbit $a(\Lambda)$ and the plane orthogonal to $\theta$ through orbit point $a(\lambda)$. The term $\frac{1}{n(\theta, \lambda)}$ and $\mid a'(\lambda)\cdot\theta\mid$ are closely related to the orbit's geometry and could be regarded as a redundancy weight. In the next section \ref{sec:D}, we will discuss the parameters obtained through the training of the neural network to approximate this discrete redundancy weight.

The algorithm comprises three main steps. Firstly, for each cone-beam projection $\lambda\in \Lambda$, Grangeat's intermediate function is computed using~\eqref{eq:myequation2}. Secondly,  shift-variant filtering to the obtained intermediate function $S(\theta, \lambda )$ is applied: 
\begin{align*}
g^F(\omega, \lambda)=& \int_{\substack{S^2/2}} d\theta-\frac{1}{4 \pi^2} \mid a'(\lambda)\cdot\theta\mid\frac{1}{n(\theta, \lambda)}\delta'(\omega, \theta)S(\theta,\lambda) \\
&\omega\in S^2, \lambda\in \Lambda.\tag{7}
\label{eq:myequation4}
\end{align*}

Finally, the filtered cone-beam projections are backprojected into the 3D volume.
\begin{align*}
f(x)=&\int_{\substack{\Lambda}} d\lambda\frac{1}{\mid x-a(\lambda)\mid^2}g^F\left(\frac{x-a(\lambda)}{\mid x-a(\lambda)\mid}, \lambda\right) \\
&x\in \mathbb{R}^3, \mid x\mid\leq B\tag{8}
\label{eq:myequation5}
\end{align*}

\subsection{Defrise and Clack Neural Network}
\label{sec:D}
The objective of this study is to create an end-to-end neural network framework based on Defrise and Clack's algorithm \cite{Defrise}. The framework's data-driven nature enables it to learn autonomously from projection data of any specific CT orbits. This involves determining the appropriate redundancy weight for a specific CT orbit to facilitate effective image reconstruction.
\subsubsection{Grangeat's intermediate function}


A flat-panel detector is used to collect cone-beam projection data and each straight line in the detector is parametrized as usual by a radial coordinate $s$ and $\mu$.
Equation~\eqref{eq:myequation6} can then be derived from~\eqref{eq:myequation2} \cite{Defrise}.


\begin{align*}
S(s,\mu,\lambda)=\frac{s^2+D^2}{D^2}\int_{+e}^{-e}dv\frac{\partial }{\partial u}\left\{\frac{Dg(x,y,\lambda)}{\sqrt{u^2+v^2+D^2}}\right\} _{u=s}\tag{9}
\label{eq:myequation6}
\end{align*}

In this equation, $u$ and $v$ represent coordinate transformations defined as $u=x\cos\mu+y\sin \mu$ and $v=-x\sin\mu+y\cos \mu$, while $D$ denotes the distance between the source and the detector. In addition, the variable $e$ represents the radius of the region defined by the cone-beam projection of the field of view.

According to~\eqref{eq:myequation6}, the computation of Grangeat's intermediate function can be broken down into three steps, each represented by one or two layers in a neural network. The first step applies cosine weight $W_{cos}$ to the projection data. Next, we perform a 2D Radon transform $A_{2d}$ using parallel beam geometry to re-project the weighted projection data. Finally, we differentiate $D$ the sinogram obtained from the 2D Radon transform with respect to $s$ and apply weight $W_{sino}$ in the sinogram domain. 
By combining the previous steps, we formulated~\eqref{eq:myequation7}. 
\begin{align*}
S(s,\mu,\lambda)=W_{sino}DA_{2d}W_{cos}p\tag{10}
\label{eq:myequation7}
\end{align*}

\begin{figure}[!t]
\centering
\includegraphics[width=3.5in]{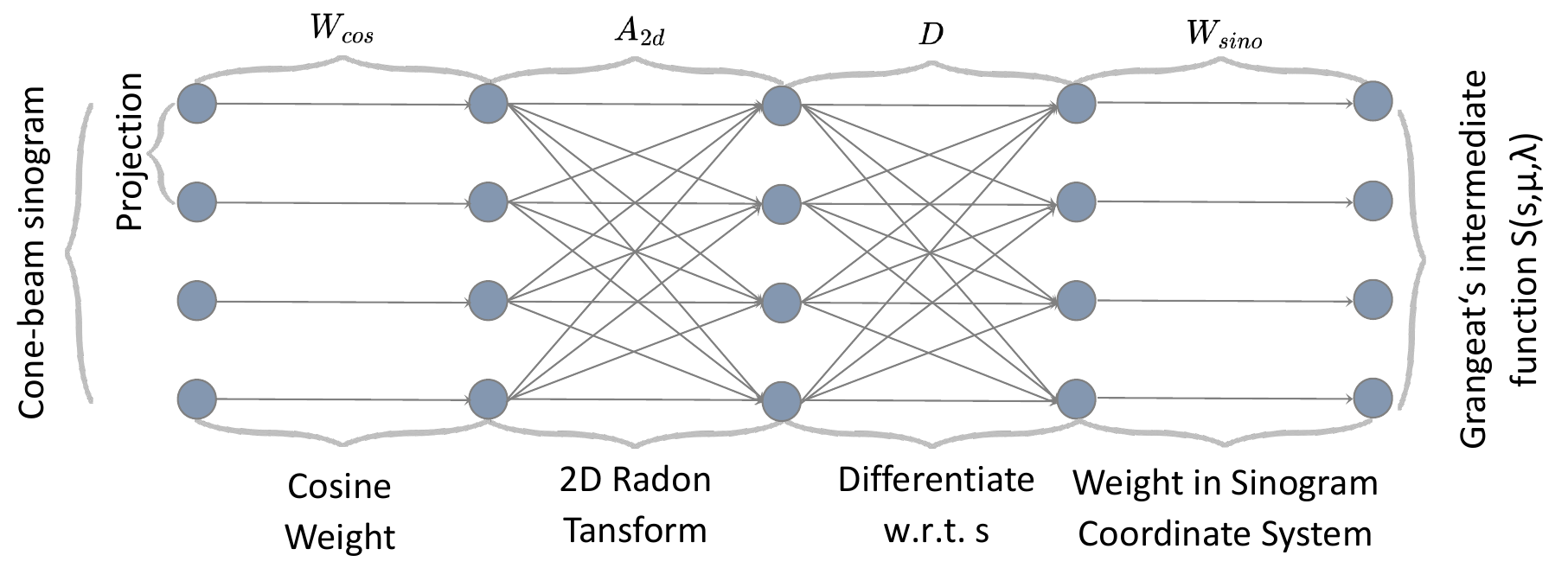}
\caption{Grangeat's intermediate function part of the neural network.}
\label{network1}
\end{figure}

This equation differs from~\eqref{eq:myequation6} mainly in the order of integration and differentiation being switched. The purpose of this switch is to reduce the noise caused by the differentiation process. Using~\eqref{eq:myequation7}, we construct the neural network shown in Fig. \ref{network1}.
\subsubsection{Filter the cone-beam data}
Equation~\eqref{eq:myequation8} is derived from~\eqref{eq:myequation4} by Defrise and Clack \cite{Defrise}.
\begin{align*}
g^F(x,y,\lambda)&=(x^2+y^2+D^2)\int_{0}^{\pi} d\mu\left\{\frac{\partial }{\partial s}\frac{S_1(s,\mu,\lambda)}{\sqrt{s^2+D^2}}\right\} \\
s&=x\cos\mu+y\sin\mu \\ 
S_1(s,\mu,\lambda)&=-\frac{1}{4\pi^2}\mid a'(\lambda)\cdot\theta\mid M(\theta,\lambda)S(\theta,\lambda)\tag{11}
\label{eq:myequation8}
\end{align*}

A neural network is constructed based on~\eqref{eq:myequation8}. The redundancy weight $W_{red}$ applied to the projection data is closely related to the orbital geometry and is set as learnable parameters. The weighted sinogram is then differentiated $D$ with respect to s in the sinogram domain. After that, the filtered sinogram is backprojected $A_{2d}^T$ into the detector coordinate system using the same parallel beam geometry. Finally, weight $W_d$ is applied in the detector domain for geometric correction. Following these steps will result in~\eqref{eq:myequation9}.

\begin{align*}
g^F(x,y,\lambda)=W_{d}A_{2d}^TDW_{red}S(s,\mu,\lambda)\tag{12}
\label{eq:myequation9}
\end{align*}

\begin{figure}[!t]
\centering
\includegraphics[width=3.5in]{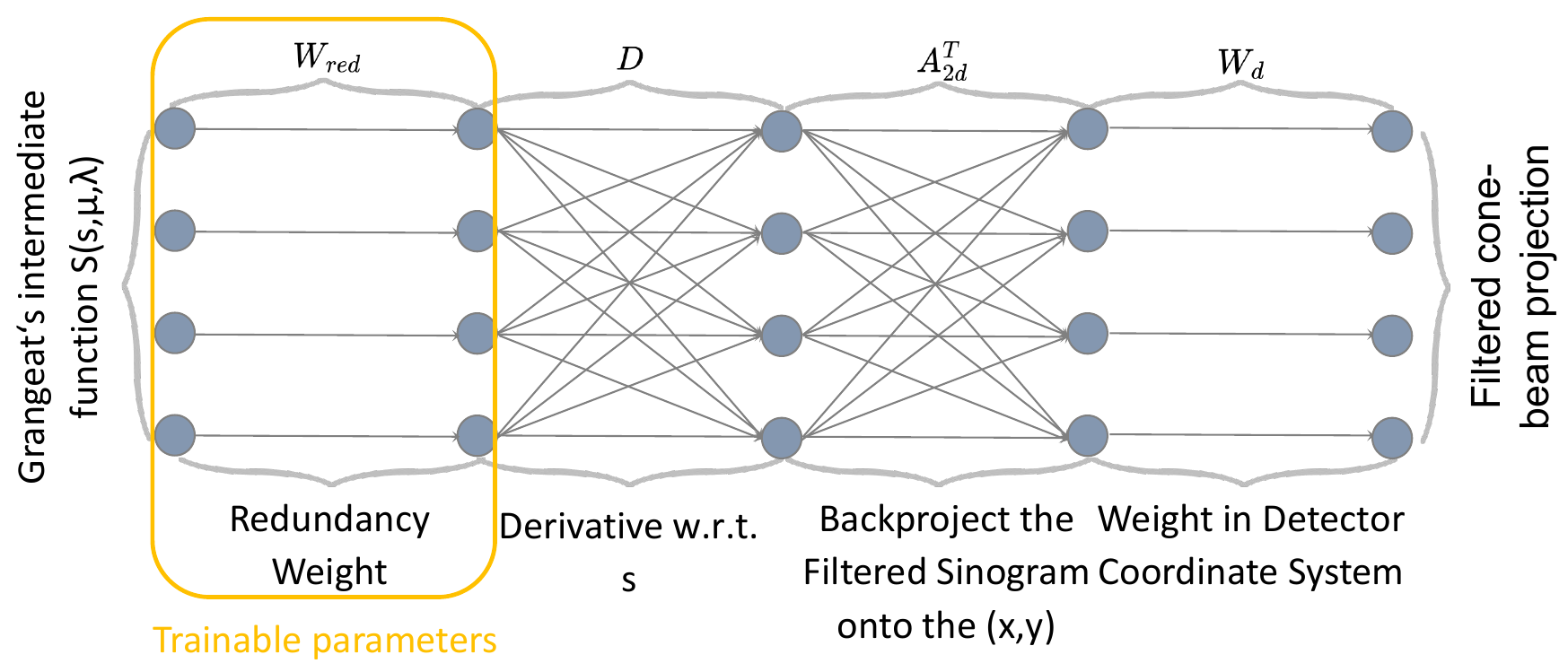}
\caption{Filtering part of the neural network.}
\label{network2}
\end{figure}

The filtering part of the neural network is illustrated in Fig. \ref{network2}.

\subsubsection{Backprojection to the 3D image}
\begin{figure*}[!t]
\centering
\subfloat{\includegraphics[width=3.7in]{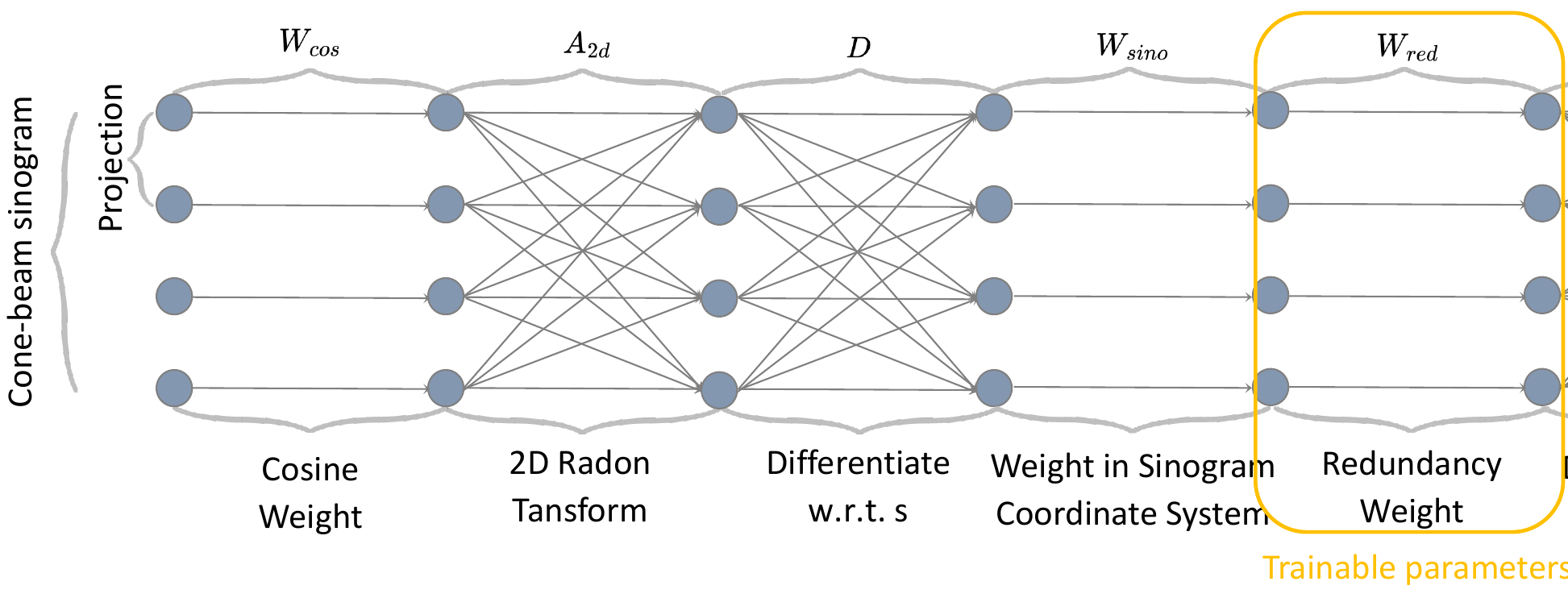}%
\label{fig1_first_case}}
\subfloat{\includegraphics[width=3.7in]{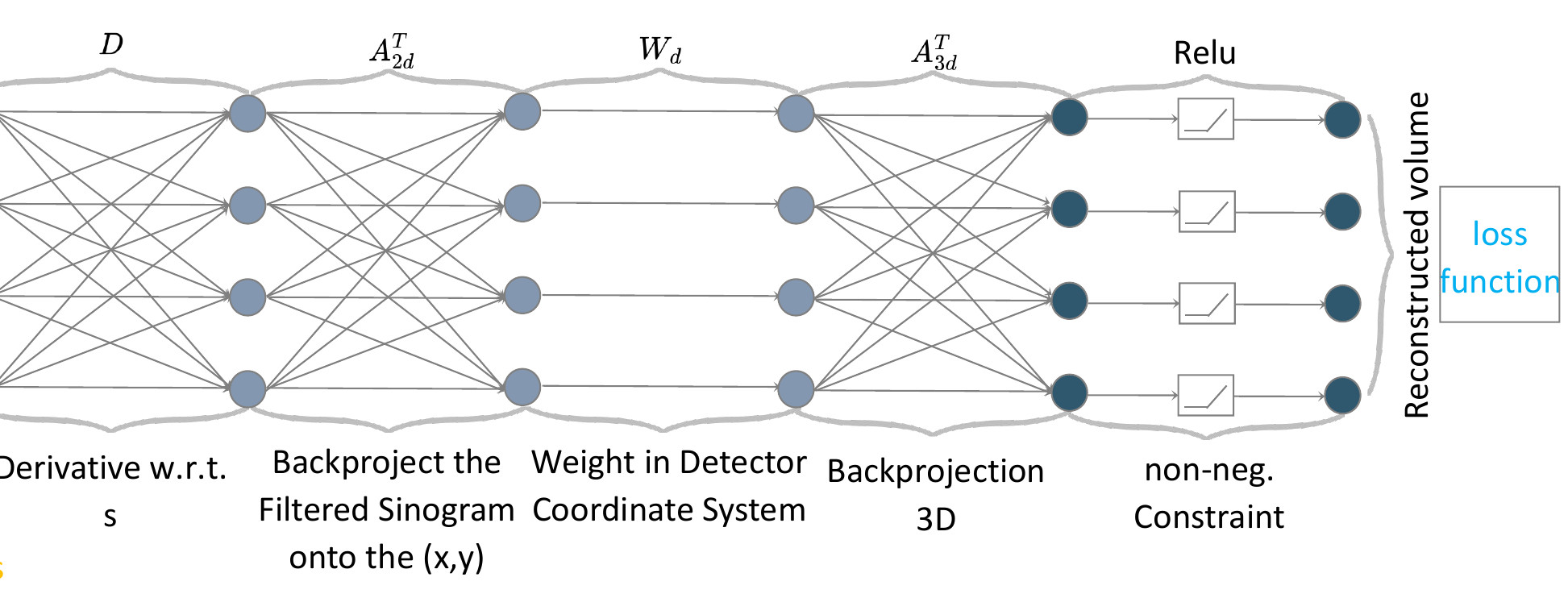}%
\label{fig1_second_case}}
\caption{Grangeat's intermediate function part, filtering part, and backprojection part are combined to form the Defrise and Clack neural network architecture.}
\label{network}
\end{figure*} 

All of the above steps can be considered as shift-variant filtering. Afterward, the filtered cone beam projection data is back-projected into a 3D volume $A_{3d}^T$. The output layer of the neural network uses ReLU to enforce non-negativity. By combining~\eqref{eq:myequation7} and~\eqref{eq:myequation9}, we obtain the complete reconstruction formula~\eqref{eq:myequation10}. 
\begin{align*}
x=A_{3d}^TW_{d}A_{2d}^TDW_{red}W_{sino}DA_{2d}W_{cos}p\tag{13}
\label{eq:myequation10}
\end{align*}

Based on this formula, the complete network architecture has been constructed, as shown in Fig. \ref{network}. 

\section{Experiments and Results}
\label{sec: A}

For our study, we selected the walnut dataset \cite{walnut}, an open collection of X-ray cone-beam CT datasets of forty-two walnuts. The dataset includes raw projection data, scanning geometry details, pre-processing, and reconstruction scripts, making it suitable for tasks such as image reconstruction.
We generated 30 simulated data samples to obtain a suitable training and validation dataset. Twenty-four of the sets are used for training, while the remaining sets are used for validation. For each set, we randomly generated 5 to 10 geometric objects of different types, positions, and rotation directions within a blank voxel volume as the ground truth. We followed the same orbit geometry as the walnut dataset and used the forward projection of the software PyroNN \cite{syben2019pyro} to generate the sinogram, which served as input for the neural network. A subset of 10 walnuts was used as the test dataset to evaluate the model's performance on real-world data.

Table \ref{tab: parameters} presents the specific parameters of the circular orbit geometry.
\begin{table}[htbp]
\centering
\caption{Parameters of the circular orbit geometry}
\resizebox{\linewidth}{!}{%
\begin{tabular}{{@{}cccc@{}}}
\toprule
\textbf{Volume shape} & \textbf{Volume spacing} & \textbf{Number of projections} \\ 
\midrule
501$\times$501$\times$501 & 0.1mm$\times$0.1mm$\times$0.1mm & 400 \\

\toprule
\textbf{Angular range} & \textbf{Source isocenter distance} & \textbf{Source detector distance}\\
\midrule
$2\pi$ & 66mm & 199mm \\
 
\toprule
\textbf{Detector shape} & \textbf{Detector spacing}  \\
\midrule
972$\times$768 & 0.1496mm$\times$0.1496mm\\
\bottomrule
\end{tabular}%
}
\label{tab: parameters}
\end{table}


In our experiment, we used PyTorch 2.1.1 to construct the neural network's architecture. The 2D Radon transform and the 3D cone-beam backprojection were implemented using operators from the PyroNN library \cite{syben2019pyro}. We chose Mean Squared Error (MSE) as the loss function. Additionally, we utilized the Adam optimizer with an initial learning rate of $1 \times 10^{-5}$ and adjusted it within a specific range using the Onecycle learning rate policy, which enhanced the training process. Random initialization was also applied to the redundancy weight layer $W_{red}$. The model ultimately converged after undergoing training for 10 epochs on an Nvidia A40 GPU.

In our experiments, we observed unstable convergence of the network which introduced high-frequency noise. This adversely affected the reconstruction results. Therefore, we decided to apply Gaussian filtering to the learned parameters before using them for the reconstruction process.

\begin{figure}[!t]
\centering
\subfloat[]{\includegraphics[width=1.2in]{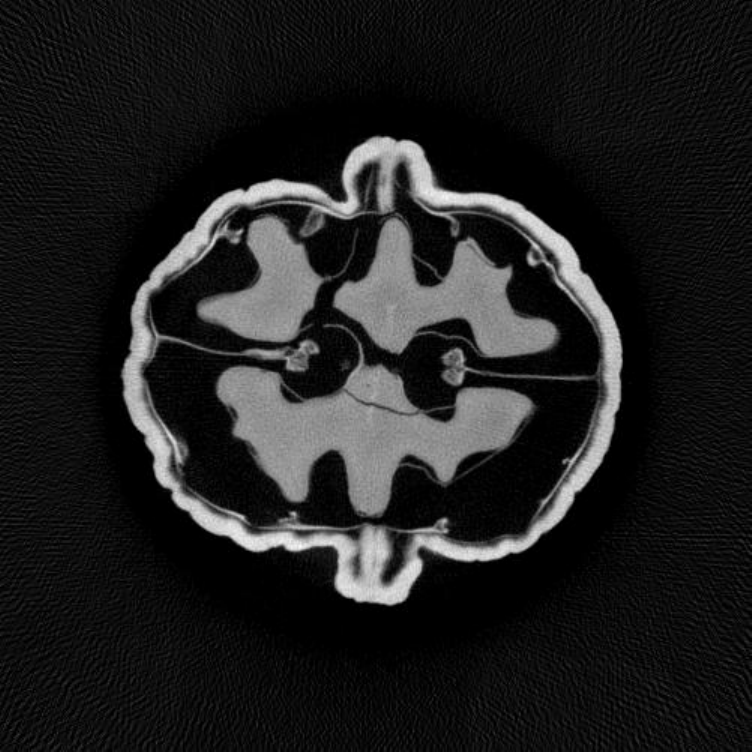}%
\label{fig_first_casea}}
\subfloat[]{\includegraphics[width=1.2in]{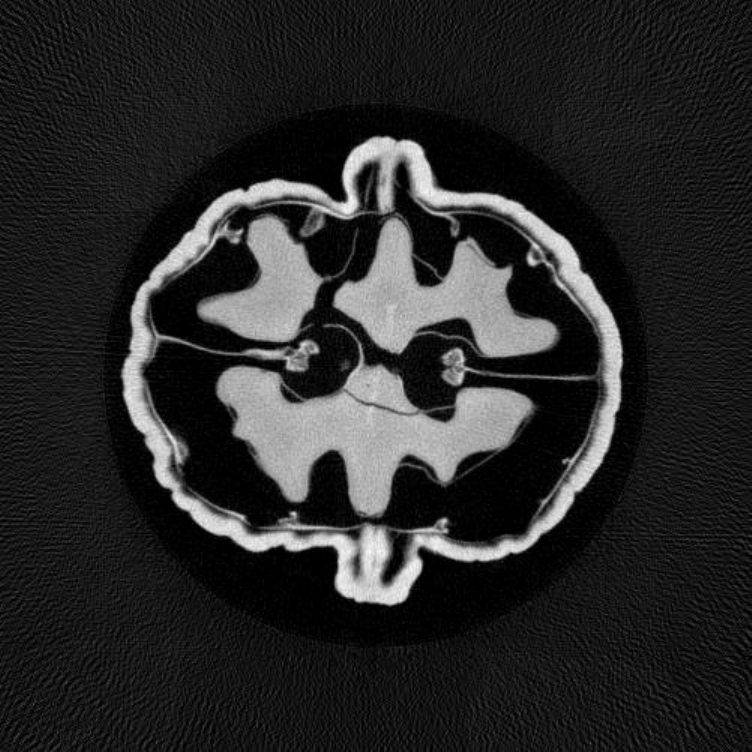}%
\label{fig_second_casea}}
\subfloat[]{\includegraphics[width=1.2in]{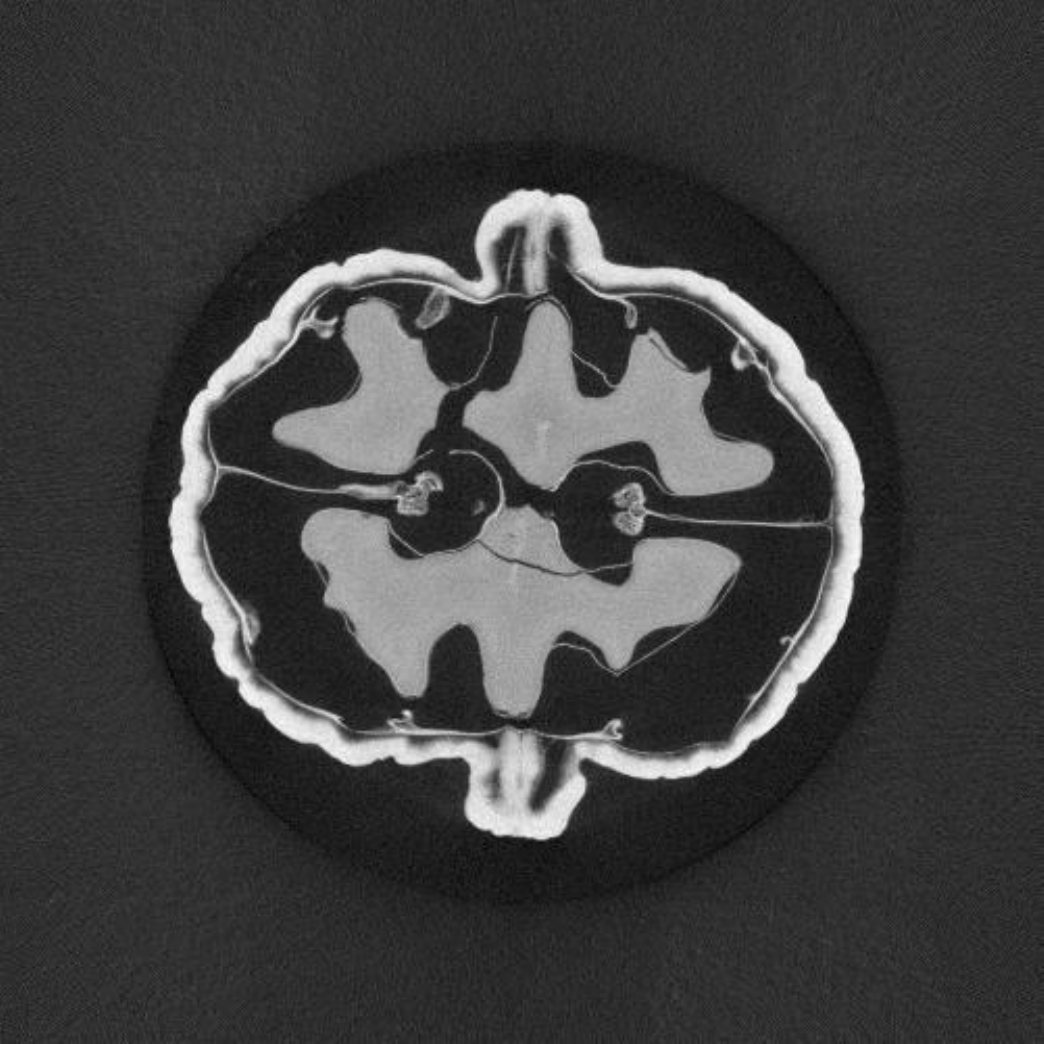}%
\label{fig_third_casea}}
\caption{Reconstructed results of the network. (a) Reconstruction using learned redundancy weight. (b) Reconstruction using analytic redundancy weight. (c) FDK reconstruction result.}
\label{reconstructed}
\end{figure}

In Fig. \ref{reconstructed}, a comparison of the central slice of volumes reconstructed with different redundancy weights is presented. The results show that the neural network, after training, can achieve reconstruction results similar to those obtained through analytical methods.

\section{Conclusion and Discussion}


Our research proposes a reconstruction pipeline for specific CBCT orbits using known operators. We have demonstrated the effectiveness of a data-driven approach for optimizing parameters in this reconstruction pipeline.  Our findings indicate that the Defrise and Clack neural network can learn parameters for reconstruction based on projection data with circular orbit geometry. 
This trained neural network performs well in reconstruction. 
For non-circular orbits, it is necessary to generate projection data specific to the given orbit geometry and subsequently optimize the parameters of the neural network to fit this orbit geometry. In summary, the Defrise and Clack neural network has the potential to reconstruct projection data with any specific orbit.
These improvements have implications for medical and industrial imaging, representing a major advancement in C-arm CT imaging technology. Future research should focus on improving reconstructed image quality, optimizing this neural network model to further speed up reconstruction, and exploring further applications.
%


%



\ifCLASSOPTIONcaptionsoff
  \newpage
\fi

\end{document}